\documentclass[letterpaper]{article} 
\usepackage{aaai2026}  
\usepackage{times}  
\usepackage{helvet}  
\usepackage{courier}  
\usepackage[hyphens]{url}  
\usepackage{graphicx} 
\usepackage{natbib}  
\usepackage{caption} 
\usepackage{algorithm}
\usepackage{algorithmic}

\usepackage{amsmath}
\usepackage{amsfonts}
\usepackage{multirow}
\usepackage{booktabs}
\usepackage{pifont}
\usepackage{makecell}

\usepackage{xcolor} 
\usepackage{colortbl}
\usepackage[many]{tcolorbox} 
\definecolor{mygray}{gray}{.9}
\usepackage{enumitem}

\newtcolorbox{promptbox}[1][]{
    breakable, 
    colback=gray!5!white,      
    colframe=gray!75!black,    
    fonttitle=\bfseries,       
    title=#1,                  
    arc=2mm,                   
    boxrule=0.5pt,             
    left=5mm, right=5mm, top=3mm, bottom=3mm, 
    before upper={
        \setlist[enumerate,1]{nosep, leftmargin=*, label=\arabic*.}
        \setlist[enumerate,2]{nosep, leftmargin=*, label=(\arabic*)}
        \setlist[itemize]{nosep, leftmargin=*}
    },
}
\newcommand{\promptrule}{\vspace{1.5ex}\hrule\vspace{1.5ex}}

\usepackage{newfloat}
\usepackage{listings}
\DeclareCaptionStyle{ruled}{labelfont=normalfont,labelsep=colon,strut=off} 
\floatstyle{ruled}
\newfloat{listing}{tb}{lst}{}
\floatname{listing}{Listing}
\title{MMhops-R1: Multimodal Multi-hop Reasoning}

\author{
    Tao Zhang\textsuperscript{\rm 1, \rm 2, \rm 3, \rm 4}, Ziqi Zhang\textsuperscript{\rm 1, \rm 3}, Zongyang Ma\textsuperscript{\rm 1, \rm 3}, Yuxin Chen\textsuperscript{\rm 4}, Bing Li\textsuperscript{\rm 1, \rm 3, \rm 5}\thanks{Corresponding author.}, Chunfeng Yuan\textsuperscript{\rm 1, \rm 3}, Guangting Wang\textsuperscript{\rm 4}, Fengyun Rao\textsuperscript{\rm 4}, Ying Shan\textsuperscript{\rm 4}, Weiming Hu\textsuperscript{\rm 1, \rm 2, \rm 3, \rm 6}
}

\affiliations{
    \textsuperscript{\rm 1}State Key Laboratory of Multimodal Artificial Intelligence Systems, CASIA,\\
    \textsuperscript{\rm 2}School of Artificial Intelligence, University of Chinese Academy of Sciences,\\
    \textsuperscript{\rm 3}Beijing Key Laboratory of Super Intelligent Security of Multi-Modal Information,\\
    \textsuperscript{\rm 4}Tencent Inc.,\\
    \textsuperscript{\rm 5}PeopleAl Inc.,\\
    \textsuperscript{\rm 6}School of Information Science and Technology, ShanghaiTech University\\
    \{zhangtao2023, mazongyang2020\}@ia.ac.cn,\\
    \{ziqi.zhang, bli, cfyuan, wmhu\}@nlpr.ia.ac.cn,\\
    \{uasonchen, guangtwang, fengyunrao, yingsshan\}@tencent.com
}

\begin{document}

\maketitle

\begin{abstract}
The ability to perform multi-modal multi-hop reasoning by iteratively integrating information across various modalities and external knowledge is critical for addressing complex real-world challenges. However, existing Multi-modal Large Language Models (MLLMs) are predominantly limited to single-step reasoning, as existing benchmarks lack the complexity needed to evaluate and drive multi-hop abilities. To bridge this gap, we introduce \textbf{MMhops}, a novel, large-scale benchmark designed to systematically evaluate and foster multi-modal multi-hop reasoning. MMhops dataset comprises two challenging task formats, \textbf{Bridging} and \textbf{Comparison}, which necessitate that models dynamically construct complex reasoning chains by integrating external knowledge. To tackle the challenges posed by MMhops, we propose \textbf{MMhops-R1}, a novel multi-modal Retrieval-Augmented Generation (mRAG) framework for dynamic reasoning. Our framework utilizes reinforcement learning to optimize the model for autonomously planning reasoning paths, formulating targeted queries, and synthesizing multi-level information. Comprehensive experiments demonstrate that MMhops-R1 significantly outperforms strong baselines on MMhops, highlighting that dynamic planning and multi-modal knowledge integration are crucial for complex reasoning. Moreover, MMhops-R1 demonstrates strong generalization to tasks requiring fixed-hop reasoning, underscoring the robustness of our dynamic planning approach. 
\end{abstract}

\begin{links}
    \link{Code}{https://github.com/taoszhang/MMhops-R1}
\end{links}

\section{Introduction}

\begin{figure}[t]
    \centering
    \includegraphics[width=\columnwidth]{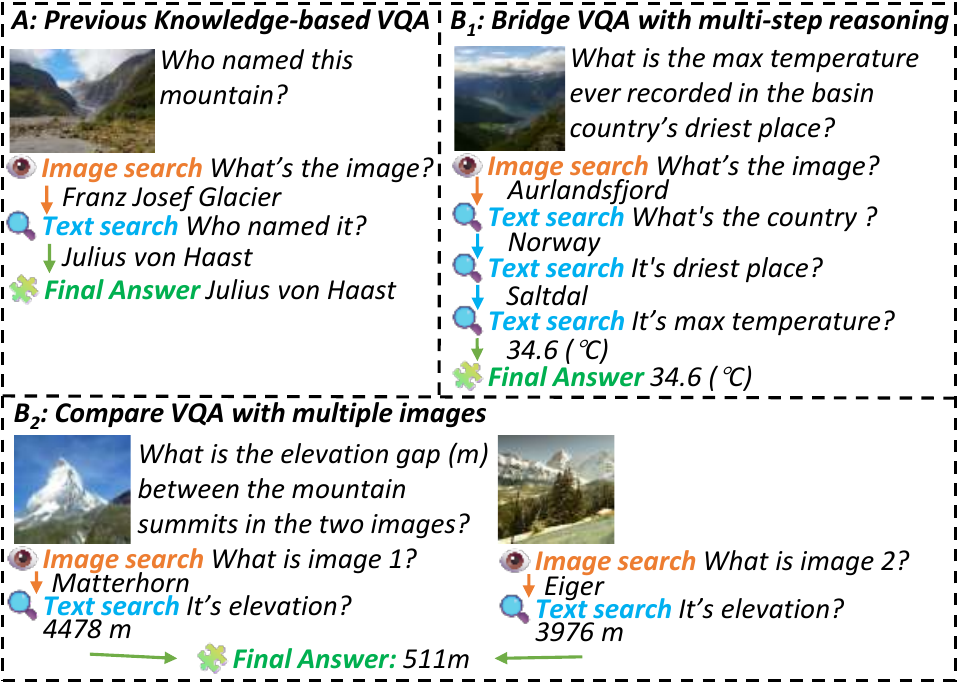}
    \caption{Comparison of reasoning types. (A) Previous KB-VQA: Single-step visual recognition followed by knowledge retrieval. (B$_{1}$) Bridging reasoning: Multi-step sequential inference on a single image. (B$_{2}$) Comparison reasoning: Cross-image entity identification and comparative analysis.}
    \label{fig1}
\end{figure}

With continuous advancement in reasoning capabilities, Large Language Models (LLMs) like OpenAI's o1 \cite{jaech2024openai}, DeepSeek-R1 \cite{guo2025deepseek}, and Kimi-k1.5 \cite{team2025kimi} demonstrate strong performance in complex problem-solving by extending chain-of-thought reasoning during inference. Multimodal large language models (MLLMs), by inheriting the reasoning abilities or adopting similar training paradigms, achieve significant progress in integrating visual understanding and language reasoning \cite{xu2024llava, peng2025skywork, zhang2025r1, zheng2025deepeyes}. However, current multimodal reasoning research primarily focuses on stimulating intrinsic model capabilities, such as spatial reasoning \cite{zhou2025r1}, object detection \cite{chen2025vinci, liu2025visual}, and mathematical reasoning \cite{meng2025mm, leng2025mmr1}. By contrast, complex real-world problems typically require the integration of multimodal reasoning with external knowledge retrieval through multi-turn interactions, enabling multimodal multi-hop reasoning. For example, Figure~\ref{fig1}(B$_{1}$) requires the model to extract information from the image, retrieve relevant external knowledge, and perform multi-step reasoning to reach the answer. Figure~\ref{fig1}(B$_{2}$) requires identifying entities across multiple images, retrieving corresponding knowledge, and conducting quantitative reasoning.

Despite progress, existing Visual Question Answering (VQA) datasets remain insufficient for multimodal multi-hop reasoning.
Current datasets are limited in both visual and textual reasoning depth: standard VQA datasets typically require only single-step visual understanding \cite{goyal2017making, hudson2019gqa, singh2019towards}, while several knowledge-based VQA datasets \cite{marino2019ok, schwenk2022okvqa, lerner2022viquae, chen2023can} introduce external knowledge retrieval to increase complexity. However, as shown in Figure~\ref{fig1}(A), models usually use one step of visual recognition and one step of text retrieval to answer, without constructing complex reasoning chains. E-VQA \cite{mensink2023encyclopedic} extends questions to two-hop reasoning, but this extension remains restricted to the textual domain and features a fixed reasoning path length, lacking multimodal integration and diverse reasoning types. These limitations make existing datasets inadequate for effectively supporting model training and evaluation in complex multimodal multi-hop reasoning tasks.

Based on these challenges, we propose \textbf{MMhops}, a novel large-scale Multimodal Multi-hop reasoning dataset that systematically increases reasoning depth in both visual and textual dimensions. MMhops features two types of reasoning tasks: \textbf{Bridging} reasoning and \textbf{Comparison} reasoning. Bridging reasoning starts from a single image and requires the model to perform multi-step chain reasoning, with each step building on the previous one, supporting reasoning depths from two hops and beyond. Comparative reasoning is based on multiple images, requiring the model to identify multiple visual entities and compare their shared attributes, involving cross-image information integration and comparative analysis. Both task types demand deep reasoning abilities in visual understanding and textual inference, enabling the model to decompose complex questions and dynamically construct answers through multi-round interactions, thus providing a comprehensive evaluation of multimodal reasoning and knowledge integration capabilities.

To address the challenges of multimodal multi-hop reasoning, we propose \textbf{MMhops-R1}, the first framework to leverage reinforcement learning (RL) for multimodal multi-hop reasoning. MMhops-R1 adopts a dynamic interaction strategy that overcomes the limitations of fixed processes in conventional multimodal Retrieval-Augmented Generation (mRAG) frameworks. Specifically, the model supports three core actions: \textbf{1)} selecting an input image and invoking the image retriever; \textbf{2)} submitting a text query to the text retriever; and \textbf{3)} generating answer based on the current information. With a tailored reward mechanism, MMhops-R1 can autonomously select reasoning strategies, dynamically adjust reasoning depth according to question complexity, and adaptively plan the reasoning path. 

We evaluate MMhops-R1 on the proposed MMhops benchmark against four categories of strong baselines: open-source MLLMs, multi-hop RAG, multimodal RAG, and proprietary MLLMs. Results demonstrate the profound effectiveness of our proposed RL-driven framework for dynamic mRAG and underscores two critical requirements for complex multi-modal reasoning: the ability to integrate multimodal external knowledge and to dynamically interact with a retrieval system. Furthermore, MMhops-R1 shows strong generalization, achieving robust performance on the single-hop questions from INFOSEEK and the two-hop questions from E-VQA. These findings validate our contributions and highlight the potential of our approach to drive future research in multi-modal multi-hop reasoning.

Our contributions are summarized as follows:
\begin{itemize}
    \item We introduce \textbf{MMhops}, the first large-scale benchmark for multimodal multi-hop reasoning, requiring the synthesis of diverse visual and textual information across various reasoning depths.
    \item We propose \textbf{MMhops-R1}, a novel mRAG framework that leverages reinforcement learning to optimize the model, enabling it to dynamically interact with multiple retrievers and adaptively plan the reasoning path.
    \item We set a new state-of-the-art on multimodal multi-hop reasoning tasks, demonstrating the superiority of our dynamic mRAG framework over existing methods.
\end{itemize}

\begin{figure*}[th]
    \centering
    \includegraphics[width=\textwidth]{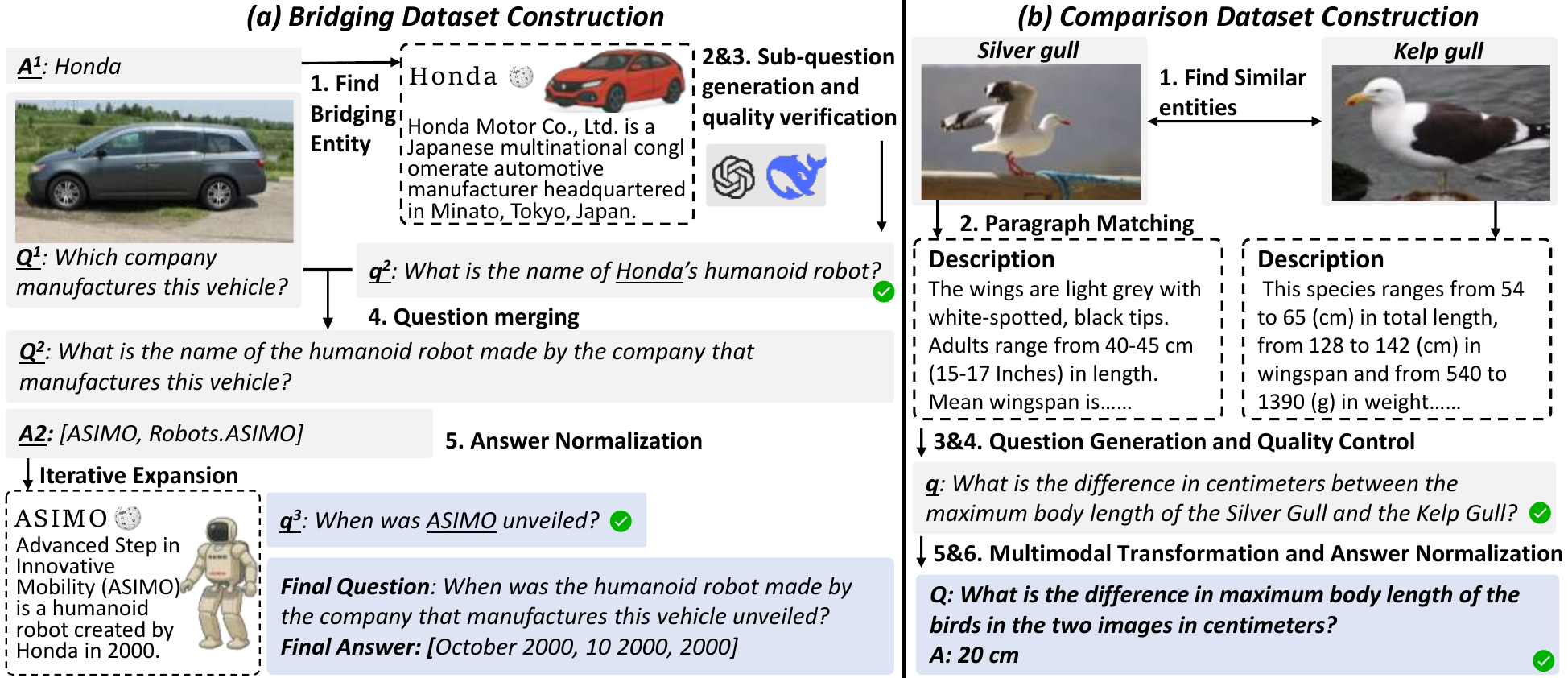}
    \caption{The multi-stage construction process for the MMhops dataset.}
    \label{fig2}
\end{figure*}

\section{Related Work}

\textbf{Knowledge-Based VQA.}
To advance VQA beyond perception towards more complex reasoning, the task of KB-VQA was introduced, which requires models to incorporate external knowledge.
However, prominent KB-VQA datasets like OK-VQA~\cite{marino2019ok} and A-OKVQA~\cite{schwenk2022okvqa} were largely confined to commonsense knowledge or simple facts.
Subsequent efforts, including ViQuAE~\cite{lerner2022viquae} and INFOSEEK~\cite{chen2023can}, expanded the knowledge domain to large-scale corpora such as Wikipedia.
Nonetheless, these datasets predominantly feature questions solvable via a two-step process: identifying a visual entity and executing a single query against a knowledge base.
While the recent E-VQA~\cite{mensink2023encyclopedic} dataset introduced textual two-hop reasoning, its reasoning chains are confined to the textual modality and a fixed length.
In contrast, the MMhops dataset is the first to systematically require multi-hop reasoning across both visual and textual modalities, featuring diverse, variable-length reasoning paths, demanding a more profound integration of multi-modal information.

\noindent \textbf{Multimodal and Multi-hop RAG.}
Early mRAG frameworks~\cite{caffagni2024wiki, yan2024echosight, zhang2024mr} typically employ a static, single-step pipeline: they first retrieve relevant documents based on the initial query and then feed them to the generator.
A key limitation of these approaches is their reliance on a static, pre-defined process, which lacks the flexibility to adapt to queries of varying complexity.
While recent work like OmniSearch~\cite{comanici2025gemini} introduces a planning agent, it relies on manually engineered prompts or supervised fine-tuning, which does not equip the model with the intrinsic capability to learn complex reasoning policies autonomously.
In parallel, multi-hop RAG has emerged in the unimodal text domain to address similar challenges.
To move beyond fixed reasoning chains, methods such as Search-R1~\cite{jin2025search} and ReSearch~\cite{chen2025learning} leverage reinforcement learning (RL) with algorithms like GRPO~\cite{shao2024deepseekmath} and PPO~\cite{schulman2017proximal} to train an agent that learns a dynamic retrieval policy.
However, these powerful RL-based paradigms have thus far been confined to the textual modality.
Our work, MMhops-R1, bridges this divide by extending this RL-based paradigm to the mRAG domain, training an agent to strategically orchestrate retrieval and reasoning across both visual and textual knowledge sources.

\begin{table*}[t]
\centering
\small
\setlength{\tabcolsep}{0.8mm}
\begin{tabular}{lrccccc}
\toprule
\textbf{Dataset} & \textbf{Scale} & \textbf{Visual Reasoning} & \textbf{Text Reasoning} & \textbf{Total Reasoning} & \textbf{Multi-image} & \textbf{Knowledge Source} \\
\midrule
OK-VQA \cite{marino2019ok} & 14K & 1 & 1 & 2 & \ding{55} &  Factoid \\
A-OKVQA \cite{schwenk2022okvqa} & 24.9K & 1 & 1 & 2 & \ding{55} &  Common sense/Factoid\\
ViQuAE  \cite{lerner2022viquae} & 3.7K & 1 & 1 & 2 & \ding{55} &  Wikipedia \\
INFOSEEK \cite{chen2023can} & 1.35M & 1 & 1 & 2 & \ding{55} & Wikipedia  \\
E-VQA \cite{mensink2023encyclopedic} & 1M & 1 & 1-2 & 2-3 & \ding{55} & Wikipedia  \\
\midrule
\textbf{MMhops} & \textbf{31.1K} & \textbf{1-2} & \textbf{2-3} & \textbf{3-4} & \textbf{\ding{51}} & \textbf{Wikipedia} \\
\bottomrule
\end{tabular}%
\caption{Comparison with Existing Knowledge-based VQA Datasets.}
\label{tab:dataset_comparison}
\end{table*}

\begin{table}[ht]
\centering
\small
\begin{tabular}{lrr}
\toprule
\textbf{Statistic Dimension} & \textbf{Value} & \textbf{Percentage} \\
\midrule
\multicolumn{3}{l}{\textbf{Dataset Scale}} \\
Total VQA Samples & 31,117 & 100.0\% \\
Bridging VQA Samples & 26,437 & 85.0\% \\
Comparison VQA Samples & 4,680 & 15.0\% \\
Number of Questions Involved & 20,483 & - \\
Number of Entities Involved & 8,832 & - \\
Number of Images Involved & 28,256 & - \\
\midrule
\multicolumn{3}{l}{\textbf{Reasoning Complexity}} \\
Requiring External Knowledge & 31,117 & 100.0\% \\
3 steps & 22,016 & 70.8\% \\
4 steps & 9,101 & 29.2\% \\
\midrule
\multicolumn{3}{l}{\textbf{Content Characteristics}} \\
Average Question Length (words) & 17.3 & - \\
Average Answer Length (words) & 1.6 & - \\
\midrule
\multicolumn{3}{l}{\textbf{Answer Type Distribution}} \\
Entity-type Answers & 5,923 & 19.0\% \\
Temporal Answers & 5,016 & 16.1\% \\
Numerical Answers & 20,178 & 64.9\% \\
\bottomrule
\end{tabular}%
\caption{Statistics of MMhops Dataset}
\label{tab:mm_hop_stats}
\end{table}

\section{MMhops Dataset}
In this section, we present MMhops, a large-scale multimodal multi-hop reasoning dataset. MMhops requires models to: (1) interact with diverse external knowledge sources for targeted retrieval; (2) perform multi-step reasoning by dynamically integrating and updating knowledge across multiple retrieval and reasoning steps; (3) align and combine information from multiple images for cross-image and cross-modal reasoning.  Through a well-designed data construction and evaluation framework, MMhops serves as an important resource for multimodal reasoning research.

\subsection{MMhops Construction}

The MMhops dataset is built based on the Wikipedia Knowledge Base (KB), with an automated data annotation and quality filtering process designed using powerful language models like GPT-4o \cite{hurst2024gpt}. The dataset includes two core reasoning types: Bridging Questions and Comparison Questions, covering various reasoning depths and different numbers of image inputs.

\subsubsection{Bridging Dataset}
Bridging Questions begin with the visual information from a single image and progressively link relevant entities and knowledge through multi-step chain reasoning. We generate them iteratively by starting with single-hop questions and progressively increasing the reasoning depth, a process depicted in Figure~\ref{fig2}~(a).

\noindent \textbf{Initialization Phase:}
\begin{enumerate}
\setcounter{enumi}{-1}
    \item \textbf{Data Collection:} Gather existing single-hop knowledge-based datasets $(V, Q^{1}, A^{1})$ as the foundation for constructing multi-hop reasoning chains.
\end{enumerate}
\noindent \textbf{Iterative Expansion Phase:}
\begin{enumerate}
    \item \textbf{Bridging Entity Identification:} From the current $n \ (\geq 1)$ hop dataset $(V, Q^n, A^n)$, select samples where the answer corresponds to a Wikipedia entity $(V_i, Q^n_i, A^n_i)$, excluding vague entity types such as numbers or years. The answer $A^n_i$ is designated as the bridging entity for subsequent reasoning chains.
    \item \textbf{Sub-question Generation:} Using the Wikipedia page of the bridging entity $A^n_i$, prompt a large language model to generate a knowledge-based question $q^{n+1}_i$, ensuring that the entity $A^n_i$ is explicitly mentioned in the question and labeling the answer $A^{n+1}_i$.
    \item \textbf{Question Quality Control:} Verify that the sub-question $q^{n+1}_i$ meets the criteria of an independent single-hop question, meaning that removing the entity $A^n_i$ from $q^{n+1}_i$ should render the question unanswerable.
    \item \textbf{Question Merging:} Merge the sub-question $q^{n+1}_i$ with the current question $Q^n_i$, replacing the reference to the bridging entity $A^n_i$ in $q^{n+1}_i$ with the current question $Q^n_i$, resulting in the complete $(n+1)$ hop question $Q^{n+1}_i$.
    \item \textbf{Answer Normalization:} Categorize the answer $A^{n+1}_i$ into three types: numerical values, time-related entities (e.g., years, dates), and strings. Construct standardized answer sets for each category.
\end{enumerate}
Through this iterative process, we systematically develop multi-level reasoning question sets, ranging from two-hop to three-hop. 

\subsubsection{Comparison Dataset}
This question type evaluates cross-image reasoning, as exemplified in Figure~\ref{fig2}~(b). Generating these questions requires identifying entities across multiple images and utilizing external knowledge to formulate a comparative query. The construction process is as follows:
\begin{enumerate}
    \item \textbf{Entity Collection:} Collect a large number of visual entities from the Wikipedia knowledge base. Use the embedding model NV-Embed-v2 \cite{lee2024nv} to perform semantic similarity matching based on the entity names and summary, selecting entities with high relevance. Use LLMs to perform semantic deduplication and remove pairs of entities that refer to the same concept.
    \item \textbf{Paragraph Matching:} For the selected similar entity pairs, use a rule-based method to extract paragraphs with the same title that contain quantifiable numerical information, providing consistent background knowledge for subsequent comparative analysis.
    \item \textbf{Question Generation:} Based on the entity pairs and their background knowledge paragraphs, prompt the LLMs to focus on quantifiable attributes and generate questions that compare the attributes of the two entities.
    \item \textbf{Quality Control:} Perform automated validation to ensure that the questions are clearly stated and the answers are quantifiable and verifiable.
    \item \textbf{Multimodal Transformation:} Replace the entity names in the questions with corresponding images to construct a multimodal reasoning scenario that forces the model to reason based on visual content.
    \item \textbf{Answer Normalization:} Standardize numerical answers, ensuring that clear units are included in the question to support accurate evaluation.
\end{enumerate}

\begin{figure*}[t]
    \centering
    \includegraphics[width=\textwidth]{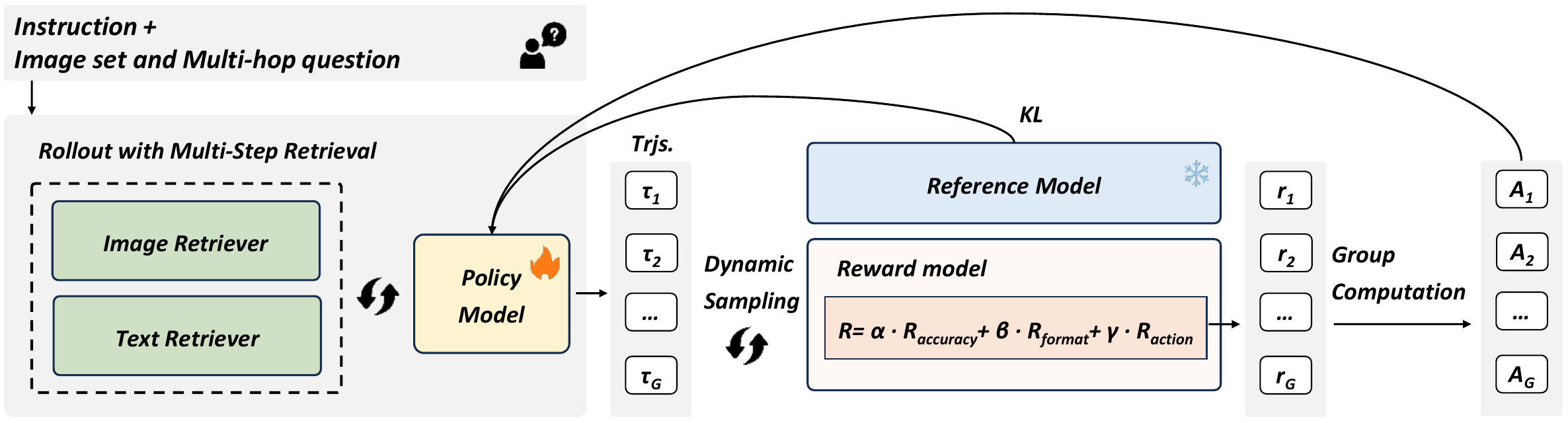}
    \caption{Overview of the training pipeline for MMhops-R1.}
    \label{fig:3}
\end{figure*}

\subsection{MMhops Analysis}
MMhops is the first large-scale dataset designed for multimodal multi-hop reasoning.
As detailed in Table~\ref{tab:mm_hop_stats}, the dataset comprises 31,117 samples, which include 20,483 unique questions, 8,832 distinct entities, and 28,256 images. 
A key feature of MMhops is its focus on complex reasoning chains; all samples require more than two reasoning hops that span both visual and textual modalities.
Specifically, 70.8\% of samples require three reasoning steps and 29.2\% require four. 
Furthermore, all samples necessitate the integration of external knowledge.
Linguistically, the average question length is 17.3 words, with concise answers averaging 1.6 words. Most answers are numerical, facilitating precise evaluation of the model's reasoning ability.

As detailed in Table~\ref{tab:dataset_comparison}, existing Knowledge-based VQA (KVQA) datasets are largely confined to shallow reasoning, typically involving a single visual step and 1--2 textual reasoning hops. 
Consequently, they are insufficient for evaluating complex, multi-step reasoning abilities.
MMhops dataset incorporates multi-image inputs, which necessitates 1--2 steps of cross-image relational reasoning.
Furthermore, we extend the textual reasoning depth to 2--3 hops via a scalable, iterative pipeline.
Collectively, these enhancements result in a total reasoning depth of 3--4 steps, establishing MMhops as a more challenging and practical benchmark to drive progress in advanced multimodal reasoning.

\subsection{Dataset Splits}
We split the MMhops dataset into training, validation, and test sets with a $7{:}1{:}2$ ratio using stratified sampling based on reasoning depth and question type.

\section{Methodology}
\subsection{Problem Formulation}
We consider the task of answering a question $Q$ based on a collection of images $\mathcal{I} = \{I_1, \ldots, I_n\}$. The policy model $\pi_{\theta}$ can leverage a set of external retrievers $\mathcal{R} = \{R_I, R_T\}$. $R_I$ is an image retriever that, given a query image, returns the information about the most similar image. $R_T$ is a text retriever that, given a text query, returns the top-$k$ most relevant passages. The model's action space is defined as $\mathcal{A} = \{a_{t}, a_{is}, a_{ts}, a_{a}\}$, where $a_t$ represents thinking and reasoning based on all current inputs, $a_{is}$ and $a_{ts}$ invoke the retrievers $R_I$ and $R_T$, respectively, and $a_a$ terminates the process by generating the final response. The training pipeline for our policy model is illustrated in Figure~\ref{fig:3}.

\subsection{Rollout with Multi-Step Retrieval}
At each time step $t$, the policy model $\pi_{\theta}$ first performs thinking based on the current state $s_t$ (containing historical interaction information), then selects the next action from the action set $\{a_{is}, a_{ts}, a_a\}$. The specific action execution mechanisms are as follows:
\begin{itemize}
    \item \textbf{Image Retrieval ($a_{is}$)}: The policy issues a query to an image retriever $R_I$ by generating target image indices, formatted within \texttt{<image\_search>} and \texttt{</image\_search>} tags. The retriever returns the corresponding image information as observation $o_{t+1}$.
    \item \textbf{Text Retrieval ($a_{ts}$)}: The policy generates a textual query, formatted within  \texttt{<text\_search>} and \texttt{</text\_search>} tags, for a text retriever $R_T$, which returns the top-$k$ relevant passages as observation $o_{t+1}$.
    \item \textbf{Answer ($a_a$)}: The policy generates the final answer, formatted within \texttt{<answer>} and \texttt{</answer>} tags, based on the information gathered throughout the trajectory. This is a terminal action that concludes the episode.
\end{itemize}
If the policy generates an action with a malformed syntax or one outside $\mathcal{A}$, the environment provides a fixed penalty signal as the observation $o_{t+1}$ to encourage valid action generation.
The rollout process terminates when the policy executes the answer action $a_a$ or a maximum of $T$ steps is reached. This interaction generates a trajectory $\tau$, defined as a sequence of states, actions, and observations:
\begin{equation}
   \tau = \{(s_0, a_0, o_1), (s_1, a_1, o_2), \ldots, (s_T, a_T)\}.
\end{equation}

\subsection{Reward Modeling}
To guide the model's generation, we design a composite reward function for MMhops-R1. This function comprises three components designed to promote correctness, structural clarity, and effective tool use. For a given trajectory $\tau$, the total reward is a weighted sum of these components.

\begin{enumerate}
    \item \textbf{Outcome Reward ($R_{\text{outcome}}$).}
    This binary reward evaluates the correctness of the final answer. It is defined as $R_{\text{outcome}}(\tau) = 1$ if the model's answer matches the ground truth, and $0$ otherwise.

    \item \textbf{Format Reward ($R_{\text{format}}$).}
    This binary reward encourages adherence to the previously defined structured format. A trajectory receives $R_{\text{format}}(\tau) = 1$ if all generated thoughts and actions are correctly formatted with their respective tags, and $0$ otherwise.

    \item \textbf{Action Reward ($R_{\text{action}}$).}
    This rewards effective tool use. A key aspect of our design is that this reward is gated by the overall success of the trajectory. It is only granted if the model both produces the correct final answer and adheres to the required format. This encourages the model to learn tool-use policies that directly contribute to successful outcomes. The reward is defined as:
    \begin{equation}
        R_{\text{action}}(\tau) = R_{\text{outcome}}(\tau) \cdot R_{\text{format}}(\tau) \cdot R_{\text{tool}}(\tau)
    \end{equation}
    where $R_{\text{tool}}(\tau)$ is a separate reward, defined as the number of syntactically correct tool invocations.
\end{enumerate}

The total reward for a trajectory $\tau$ is a weighted sum of these components:
\begin{equation}
    R(\tau) = \alpha \cdot R_{\text{outcome}}(\tau) + \beta \cdot R_{\text{format}}(\tau) + \gamma \cdot R_{\text{action}}(\tau)
    \label{eq:total_reward}
\end{equation}
where $\alpha, \beta, \text{and } \gamma$ are non-negative hyperparameters that balance the contribution of each component.

\begin{table*}[t]
\centering
\small
\setlength{\tabcolsep}{1.0mm}
\begin{tabular}{lllccccc}
\toprule
\multicolumn{1}{c}{} & \multicolumn{1}{c}{} & & \multicolumn{4}{c}{\textbf{Bridging}}  &   \\ \cmidrule(l){4-7} 
\multicolumn{1}{c}{\multirow{-2}{*}{\textbf{Method}}} & \multicolumn{1}{c}{\multirow{-2}{*}{\textbf{Base Model}}} & \multirow{-2}{*}{\textbf{Retriever}} & \textbf{String} & \textbf{Numerical} & \textbf{Time}  & \textbf{Overall} & \multirow{-2}{*}{\textbf{Comparison}} \\ \midrule
\multicolumn{8}{l}{\textbf{Closed-sourced model}}  \\
GPT-4o-mini \cite{hurst2024gpt}  & -- & --  & 29.67  & 20.54   & 26.08 & 23.80   & 7.05     \\
GPT-4o \cite{hurst2024gpt} & --  & --  & 41.66  & 33.60   & 39.28 & 36.62   & 8.76    \\
Gemini-2.5-flash \cite{comanici2025gemini} & -- & -- & 51.08  & 43.51 & 50.10 & 46.58 & 23.18 \\
 Gemini-2.5-pro \cite{comanici2025gemini} & -- & -- & \textbf{58.80}  & \textbf{50.83} & \textbf{57.32} & \textbf{53.98} & \textbf{29.39} \\ \midrule
\multicolumn{8}{l}{\textbf{Direct Answer}}  \\
Zero-shot & Qwen2.5-vl-7B-Instruct  & -- & 24.21  & 15.24 & 26.60 & 19.53   & 6.20 \\
Zero-shot & Qwen2.5-vl-72B-Instruct & -- & 37.51  & 32.11 & 37.32   & 34.39   & 7.59   \\
\midrule
\multicolumn{8}{l}{\textbf{Multi-hop RAG (Text-only)}}  \\ 
Search-r1 \cite{jin2025search} & Qwen2.5-7b-Instruct & Caption, Text  & 14.45 & 23.05  &  17.85  & 19.98  & 6.62   \\
Self-Ask \cite{press2022measuring} & GPT-4o & Caption, Text  &  27.59 & 31.41 & 31.13  & 30.42 & 18.27   \\ \midrule
\multicolumn{8}{l}{\textbf{Multimodal RAG}}  \\
Vanilla mRAG & Qwen2.5-vl-7B-Instruct & Text  & 14.37  & 14.65  & 14.95 & 14.63   & 3.95     \\
Vanilla mRAG & Qwen2.5-vl-7B-Instruct & Image, Text  & 26.52  & 25.68 & 28.97 & 26.49  & 9.72 \\
EchoSight \cite{yan2024echosight} & LLaMA3 & Image, Text   &  19.14 & 11.83 & 11.86 & 13.63  & 4.81    \\
OmniSearch \cite{li2024benchmarking} & GPT-4o & Image, Text   & 31.02 & 49.77 & 36.5  & 42.65  & 17.02   \\ 
\textbf{MMhops-R1 (Ours)} & \textbf{Qwen2.5-vl-7B-Instruct} & \textbf{Image, Text}  &  \textbf{44.66} & \textbf{55.33}  & \textbf{47.94} & \textbf{51.35} & \textbf{22.01}  \\  \bottomrule
\end{tabular}%
\caption{Main results on MMhops.}
\label{tab:MMhops}
\end{table*}

\subsection{Objective Function}
To optimize our policy $\pi_\theta$ using a composite reward signal, we adapt the objective function from DAPO~\cite{yu2025dapo}.
This objective is coupled with a dynamic sampling strategy that filters generated response groups.
Specifically, we enforce that each group of $G$ responses must contain at least one factually correct sample, as stipulated by the constraint in our objective.
This design enables the policy to optimize for procedural correctness by creating non-zero advantages ($\hat{A}_{i,t}$) from process-based rewards, such as format adherence ($R_{\text{format}}$), even when the final outcome is already correct.
Our full optimization objective is formulated as:
\begin{equation}
\begin{split}
J(\theta) &= \mathbb{E}_{(q,a) \sim \mathcal{D}, \{o_i\}^G_{i=1} \sim \pi_{\theta_{\text{old}}}(\cdot \mid q; R)} \\
&\quad \left[ \frac{1}{\sum_{i=1}^{G} |o_i|} \sum_{i=1}^{G} \sum_{t=1}^{|o_i|} \min \left( r_{i,t}(\theta) \hat{A}_{i,t}, \right. \right. \\
&\quad \left. \left. \text{clip} \left( r_{i,t}(\theta), 1-\epsilon_{\text{low}}, 1+\epsilon_{\text{high}} \right) \hat{A}_{i,t} \right) \right] \\
\text{s.t.} \quad & 0 < \left| \{o_i \mid \text{is\_equivalent}(a, o_i)\} \right|.
\end{split}
\end{equation}
where
\begin{equation*}
    r_{i,t}(\theta) = \frac{\pi_\theta(o_{i,t} \mid q, o_i, \leq t ; R)}{\pi_{\theta_{\text{old}}}(o_{i,t} \mid q, o_i, \leq t ; R)}, \text{and}
\end{equation*}
\begin{equation*}
    \hat{A}_{i,t} = \frac{R_i - \text{mean}(\{R_i\}^G_{i=1})}{\text{std}(\{R_i\}^G_{i=1})}
\end{equation*}
Where $R$ represents the retriever, and the model samples while interacting with multiple retrievers $R = \{ R_{I}, R_{T}\}$.

\subsection{Loss Masking for External Observations}
The observation $o_{t+1}$ at each timestep $t$ includes tokens from external sources, such as results from the image ($R_I$) and text ($R_T$) retrievers and environmental feedback on invalid actions. Since these tokens are not generated by the policy model, we mask them from the loss computation during policy optimization. This ensures that the optimization objective is confined to the model's own generated reasoning and action tokens, improving training stability.

\section{Experiments}

\subsection{Experimental Settings}

\subsubsection{Implementation Details}
We optimize our policy using the Verl framework \cite{sheng2024hybridflow}, employing Qwen2.5-VL-7B-Instruct \cite{bai2025qwen2} as the backbone model. The model is trained for a single epoch on the MMhops dataset with a constant learning rate of $1 \times 10^{-6}$. During policy optimization, we use a batch size of 256 and a group size of 8. Our knowledge base for retrieval comprises 100K Wikipedia articles, each accompanied by an image. For image retrieval, we utilize the CLIP-ViT-L/14@336px model~\cite{radford2021learning}. For text retrieval, we employ the E5 model~\cite{wang2022text} to fetch the top-3 most relevant passages for each query. The maximum number of interaction turns with the knowledge base is set to 4 during both training and inference. The hyperparameters $\alpha$, $\beta$, and $\gamma$ for the reward function are set to 1.0, 1.0, and 0.25, respectively.

\subsubsection{Evaluation Metrics}
We adopt the evaluation protocol from INFOSEEK~\cite{chen2023can}, categorizing answers into three types: \textsc{String}, \textsc{Time}, and \textsc{Numerical}. 
For \textsc{String} answers, we report Exact Match (EM) accuracy. 
For \textsc{Time} answers, we employ EM with a tolerance of $\pm1$ year. 
For \textsc{Numerical} answers, a prediction is deemed correct if it falls within a $\pm0.1$ margin of the ground truth or achieves an Intersection-over-Union (IoU) of at least 50\%.
The overall score is the weighted average of the accuracies for each type. 
We report performance on the \textit{Test} set, with a breakdown for bridging and comparison questions.

\subsection{Comparison with SOTAs}

To enable a comprehensive comparison with existing approaches, we evaluate four categories of models: advanced open-source general-purpose multimodal large models with direct answer generation, single-modal text-only multi-hop RAG methods (which convert images into descriptions and combine them with the question as input), multimodal RAG methods, and  benchmark closed-source MLLMs.
For fairness, all compared methods share the same image and text retriever as ours.
Detailed results are presented in Table \ref{tab:MMhops}, with key findings summarized as follows:

\textbf{1. Rich domain knowledge and strong reasoning capabilities are essential for solving multimodal multi-hop problems.} 
However, general-purpose open-source MLLMs are relatively weak in both aspects, posing challenges for them to generalize to this task.
Even the 72B Qwen2.5-VL model falls short by 16.96\% and 14.42\% in overall accuracy on bridging and comparison questions, respectively, compared to our 7B-based model.

\textbf{2. Incorporating visual information is fundamental to effective multimodal reasoning.}
Text-only multi-hop RAG methods are unable to access critical visual information, making it difficult to perform appropriate knowledge retrieval and accurate reasoning for multimodal multi-hop problems.
Specifically, the state-of-the-art method Self-Ask, which significantly boosts base model performance (\textit{e.g.}, GPT-4o) on textual multi-hop tasks, even shows an overall performance drop on comparison questions in MMhops compared to GPT-4o alone (30.42\% vs. 36.62\%).

\textbf{3. Accurate multi-turn reasoning and  retrieval interactions are critical to successfully solving multimodal multi-hop problems.}
Existing multimodal RAG methods, such as those designed for KB-VQA, are tailored to single-hop tasks and lack the ability to properly decompose multimodal multi-hop questions into sequential reasoning and retrieval steps, thereby limiting their answer accuracy.
Even with the support of GPT-4o's OmniSearch, the overall accuracy on bridging and comparison questions remains 9.7\% and 4.99\% lower than ours.

\textbf{4. Closed-sourced commercial MLLMs remain the performance ceiling but still fall short of real-world applicability.}
Gemini-2.5-Pro, which has likely undergone reasoning-specific optimization and large-scale pretraining, outperforms our method but answers only about half of the bridging questions correctly, with lower accuracy on comparison questions.
This underscores that multimodal multi-hop RAG remains largely unexplored.

\begin{table}[t]
\centering
\small
\setlength{\tabcolsep}{0.5mm}
\begin{tabular}{lccc}
\toprule
\multirow{2}{*}{Model} & \multicolumn{3}{c}{INFOSEEK} \\ 
\cmidrule(lr){2-4} 
 & Unseen Q & Unseen E & Overall \\ \midrule

CLIP-PaLM \cite{chen2023can}              & 22.7            & 18.5          & 20.4    \\
CLIP-FiD  \cite{chen2023can}             & 23.3            & 19.1          & 20.9    \\
Wiki-LLaVA \cite{caffagni2024wiki}            & 30.1            & 27.8          & 28.9    \\
EchoSight  \cite{yan2024echosight}            & --              & --            & 31.3    \\ 
\textbf{MMhops-R1}                 & \textbf{33.8}            & \textbf{32.6}          & \textbf{33.2}    \\ \bottomrule
\end{tabular}%
\caption{Comparison on INFOSEEK. Q: Question, E: Entity.}
\label{tab:infoseek}
\end{table}

\begin{table}[t]
\centering
\small
\begin{tabular}{ccccc}
\toprule
Model   & PaLI & PaLM & GPT-3 & \textbf{MMhops-R1} \\ \hline
Two hop & 14.7 & 22.8 & 18.7  & \textbf{23.3}      \\ \bottomrule
\end{tabular}%
\caption{Comparison on E-VQA.}
\label{tab:enc-vqa}
\end{table}

\subsection{Cross-dataset Generalization Verification}

To verify the generalizability of the proposed method, we evaluate it on two widely used knowledge-based VQA datasets: INFOSEEK \cite{chen2023can} and E-VQA \cite{mensink2023encyclopedic}.
Results on INFOSEEK show the effectiveness of MMhops-R1 on multimodal single-hop questions, while its performance on two-hop questions in E-VQA confirms its generalization ability to multi-hop reasoning.

\subsection{Ablation Studies}

\begin{table}[t]
\centering
\small
\setlength{\tabcolsep}{0.3mm}
\begin{tabular}{lccccc}
\toprule
\multicolumn{1}{c}{}  & \multicolumn{4}{c}{\textbf{Bridging}}  &   \\ \cmidrule(l){2-5} 
\multicolumn{1}{c}{\multirow{-2}{*}{\textbf{Method}}}  & \textbf{String} & \textbf{Numerical} & \textbf{Time}  & \textbf{Overall} & \multirow{-2}{*}{\textbf{Comparison}} \\ \midrule
\textbf{MMhops-R1}  &  \textbf{44.66} & \textbf{55.33}  & \textbf{47.94} & \textbf{51.35} & \textbf{22.01} \\
w/o $R_{\text{action}}$ & 39.74  & 51.19  & 46.8  & 47.57 & 20.62  \\
w/o $R_{\text{format}}$ & 43.12   & 53.64   & 47.73   & 49.97    & 14.42   \\
w/o $R_{\text{format}}$,$R_{\text{action}}$  & 40.43  & 42.68 & 40.62  & 41.75 & 13.03 \\ \bottomrule
\end{tabular}%
\caption{Effect of $R_{\text{outcome}}$, $R_{\text{format}}$ and $R_{\text{action}}$.}
\label{tab:reward}
\end{table}

\begin{table}[t]
\centering
\small
\setlength{\tabcolsep}{1.0mm}
\begin{tabular}{cccccc}
\toprule
\multicolumn{1}{c}{}  & \multicolumn{4}{c}{\textbf{Bridging}}  &   \\ \cmidrule(l){2-5} 
\multicolumn{1}{c}{\multirow{-2}{*}{\textbf{Method}}}  & \textbf{String} & \textbf{Numerical} & \textbf{Time}  & \textbf{Overall} & \multirow{-2}{*}{\textbf{Comparison}} \\ \midrule
5 & 44.20  & \textbf{56.13}  & 49.18  & \textbf{51.92} & 20.09  \\
4 &  \textbf{44.66} & 55.33  & \textbf{47.94} & \textbf{51.35} & \textbf{22.01} \\
3 & 40.05  & 51.79 & 46.70  & 47.97 & 13.78 \\
2 & 30.75  & 46.98 & 30.13  & 39.93 & 9.83 \\ \bottomrule
\end{tabular}%
\caption{Effect of the maximum retriever interaction count.}
\label{tab:round}
\end{table}

\subsubsection{Effect of $R_{\text{outcome}}$,$R_{\text{format}}$ and $R_{\text{action}}$.}
As shown in Table \ref{tab:reward}: (1) Removing either the retrieval reward $R_{\text{action}}$ or the format reward $R_{\text{format}}$ leads to a notable performance drop, particularly on comparison questions; 
(2) Removing both $R_{\text{action}}$ and $R_{\text{format}}$ results in an even greater decline. 
These findings indicate that encouraging appropriate retrieval, enforcing correct feedback formats, and imposing strong constraints on answer precision all contribute positively to model performance.

\subsubsection{Effect of Number of Interaction Rounds.}
To demonstrate that the MMhops dataset indeed requires multi-step reasoning and RAG interaction for problem solving, we report model performance in Table \ref{tab:round} under maximum rounds constrained to 2, 3, 4, and 5.
As the number of rounds increases from 2 to 4, overall performance consistently improves, while further increasing to 5 yields no significant gains but introduces more computational overhead.
Therefore, four-step reasoning is most suitable for MMhops.

\section{Conclusion}
In this work, we introduce the first large-scale multimodal multi-hop reasoning dataset MMhops to evaluate models' capabilities in multi-turn interactive reasoning and external knowledge utilization, and extensive experiments show that existing MLLMs struggle on MMhops.
To address this, we further propose a novel reinforcement learning-based framework MMhops-R1 for multimodal reasoning and RAG interaction.
Results demonstrate that MMhops-R1 substantially outperforms existing methods by effectively leveraging reasoning and retrieval capabilities.
The code, dataset, and model weights will be open-sourced to encourage future research on the multimodal multi-hop reasoning task.

\section*{Acknowledgments}
This work was supported by the National Natural Science Foundation of China (No. U24A20331, No. 62302501), the Beijing Natural Science Foundation (No. L251005, No. L243015) and the Key Research and Development Program of Xinjiang Uyghur Autonomous Region (No. 2023B01005).

\bibliography{aaai2026}

@article{zhang2024mr,
  title={mR$^2$AG: Multimodal Retrieval-Reflection-Augmented Generation for Knowledge-Based VQA},
  author={Zhang, Tao and Zhang, Ziqi and Ma, Zongyang and Chen, Yuxin and Qi, Zhongang and Yuan, Chunfeng and Li, Bing and Pu, Junfu and Zhao, Yuxuan and Xie, Zehua and others},
  journal={arXiv preprint arXiv:2411.15041},
  year={2024}
}

@article{hurst2024gpt,
  title={Gpt-4o system card},
  author={Hurst, Aaron and Lerer, Adam and Goucher, Adam P and Perelman, Adam and Ramesh, Aditya and Clark, Aidan and Ostrow, AJ and Welihinda, Akila and Hayes, Alan and Radford, Alec and others},
  journal={arXiv preprint arXiv:2410.21276},
  year={2024}
}

@article{guo2025deepseek,
  title={Deepseek-r1: Incentivizing reasoning capability in llms via reinforcement learning},
  author={Guo, Daya and Yang, Dejian and Zhang, Haowei and Song, Junxiao and Zhang, Ruoyu and Xu, Runxin and Zhu, Qihao and Ma, Shirong and Wang, Peiyi and Bi, Xiao and others},
  journal={arXiv preprint arXiv:2501.12948},
  year={2025}
}

@article{jaech2024openai,
  title={Openai o1 system card},
  author={Jaech, Aaron and Kalai, Adam and Lerer, Adam and Richardson, Adam and El-Kishky, Ahmed and Low, Aiden and Helyar, Alec and Madry, Aleksander and Beutel, Alex and Carney, Alex and others},
  journal={arXiv preprint arXiv:2412.16720},
  year={2024}
}

@article{team2025kimi,
  title={Kimi k1. 5: Scaling reinforcement learning with llms},
  author={Team, Kimi and Du, Angang and Gao, Bofei and Xing, Bowei and Jiang, Changjiu and Chen, Cheng and Li, Cheng and Xiao, Chenjun and Du, Chenzhuang and Liao, Chonghua and others},
  journal={arXiv preprint arXiv:2501.12599},
  year={2025}
}

@article{xu2024llava,
  title={Llava-cot: Let vision language models reason step-by-step},
  author={Xu, Guowei and Jin, Peng and Wu, Ziang and Li, Hao and Song, Yibing and Sun, Lichao and Yuan, Li},
  journal={arXiv preprint arXiv:2411.10440},
  year={2024}
}

@article{peng2025skywork,
  title={Skywork r1v: Pioneering multimodal reasoning with chain-of-thought},
  author={Peng, Yi and Wang, Peiyu and Wang, Xiaokun and Wei, Yichen and Pei, Jiangbo and Qiu, Weijie and Jian, Ai and Hao, Yunzhuo and Pan, Jiachun and Xie, Tianyidan and others},
  journal={arXiv preprint arXiv:2504.05599},
  year={2025}
}

@article{meng2025mm,
  title={Mm-eureka: Exploring visual aha moment with rule-based large-scale reinforcement learning},
  author={Meng, Fanqing and Du, Lingxiao and Liu, Zongkai and Zhou, Zhixiang and Lu, Quanfeng and Fu, Daocheng and Shi, Botian and Wang, Wenhai and He, Junjun and Zhang, Kaipeng and others},
  journal={CoRR},
  year={2025}
}

@article{zhang2025r1,
  title={R1-vl: Learning to reason with multimodal large language models via step-wise group relative policy optimization},
  author={Zhang, Jingyi and Huang, Jiaxing and Yao, Huanjin and Liu, Shunyu and Zhang, Xikun and Lu, Shijian and Tao, Dacheng},
  journal={arXiv preprint arXiv:2503.12937},
  year={2025}
}

@article{zheng2025deepeyes,
  title={DeepEyes: Incentivizing" Thinking with Images" via Reinforcement Learning},
  author={Zheng, Ziwei and Yang, Michael and Hong, Jack and Zhao, Chenxiao and Xu, Guohai and Yang, Le and Shen, Chao and Yu, Xing},
  journal={arXiv preprint arXiv:2505.14362},
  year={2025}
}

@misc{leng2025mmr1,
  title={Mmr1: Advancing the frontiers of multimodal reasoning},
  author={Leng, Sicong and Wang, Jing and Li, Jiaxi and Zhang, Hao and Hu, Zhiqiang and Zhang, Boqiang and Zhang, Hang and Jiang, Yuming and Li, Xin and Zhao, Deli and others},
  year={2025}
}

@article{zhou2025r1,
  title={R1-Zero's" Aha Moment" in Visual Reasoning on a 2B Non-SFT Model},
  author={Zhou, Hengguang and Li, Xirui and Wang, Ruochen and Cheng, Minhao and Zhou, Tianyi and Hsieh, Cho-Jui},
  journal={arXiv preprint arXiv:2503.05132},
  year={2025}
}

@article{liu2025visual,
  title={Visual-rft: Visual reinforcement fine-tuning},
  author={Liu, Ziyu and Sun, Zeyi and Zang, Yuhang and Dong, Xiaoyi and Cao, Yuhang and Duan, Haodong and Lin, Dahua and Wang, Jiaqi},
  journal={arXiv preprint arXiv:2503.01785},
  year={2025}
}

@misc{chen2025vinci,
  title={Vinci. R1-v: Reinforcing super generalization ability in vision-language models with less than \$3},
  author={Chen, Liang and Li, Lei and Zhao, Haozhe and Song, Yifan},
  journal={arXiv preprint arXiv:2308.15363},
  year={2025}
}

@article{bai2025qwen2,
  title={Qwen2. 5-vl technical report},
  author={Bai, Shuai and Chen, Keqin and Liu, Xuejing and Wang, Jialin and Ge, Wenbin and Song, Sibo and Dang, Kai and Wang, Peng and Wang, Shijie and Tang, Jun and others},
  journal={arXiv preprint arXiv:2502.13923},
  year={2025}
}

@article{comanici2025gemini,
  title={Gemini 2.5: Pushing the frontier with advanced reasoning, multimodality, long context, and next generation agentic capabilities},
  author={Comanici, Gheorghe and Bieber, Eric and Schaekermann, Mike and Pasupat, Ice and Sachdeva, Noveen and Dhillon, Inderjit and Blistein, Marcel and Ram, Ori and Zhang, Dan and Rosen, Evan and others},
  journal={arXiv preprint arXiv:2507.06261},
  year={2025}
}

@article{press2022measuring,
  title={Measuring and narrowing the compositionality gap in language models},
  author={Press, Ofir and Zhang, Muru and Min, Sewon and Schmidt, Ludwig and Smith, Noah A and Lewis, Mike},
  journal={arXiv preprint arXiv:2210.03350},
  year={2022}
}

@article{jin2025search,
  title={Search-r1: Training llms to reason and leverage search engines with reinforcement learning},
  author={Jin, Bowen and Zeng, Hansi and Yue, Zhenrui and Yoon, Jinsung and Arik, Sercan and Wang, Dong and Zamani, Hamed and Han, Jiawei},
  journal={arXiv preprint arXiv:2503.09516},
  year={2025}
}

@inproceedings{goyal2017making,
  title={Making the v in vqa matter: Elevating the role of image understanding in visual question answering},
  author={Goyal, Yash and Khot, Tejas and Summers-Stay, Douglas and Batra, Dhruv and Parikh, Devi},
  booktitle={Proceedings of the IEEE conference on computer vision and pattern recognition},
  pages={6904--6913},
  year={2017}
}

@inproceedings{hudson2019gqa,
  title={Gqa: A new dataset for real-world visual reasoning and compositional question answering},
  author={Hudson, Drew A and Manning, Christopher D},
  booktitle={Proceedings of the IEEE/CVF conference on computer vision and pattern recognition},
  pages={6700--6709},
  year={2019}
}

@inproceedings{singh2019towards,
  title={Towards vqa models that can read},
  author={Singh, Amanpreet and Natarajan, Vivek and Shah, Meet and Jiang, Yu and Chen, Xinlei and Batra, Dhruv and Parikh, Devi and Rohrbach, Marcus},
  booktitle={Proceedings of the IEEE/CVF conference on computer vision and pattern recognition},
  pages={8317--8326},
  year={2019}
}

@inproceedings{marino2019ok,
  title={Ok-vqa: A visual question answering benchmark requiring external knowledge},
  author={Marino, Kenneth and Rastegari, Mohammad and Farhadi, Ali and Mottaghi, Roozbeh},
  booktitle={Proceedings of the IEEE/cvf conference on computer vision and pattern recognition},
  pages={3195--3204},
  year={2019}
}

@inproceedings{schwenk2022okvqa,
  title={A-okvqa: A benchmark for visual question answering using world knowledge},
  author={Schwenk, Dustin and Khandelwal, Apoorv and Clark, Christopher and Marino, Kenneth and Mottaghi, Roozbeh},
  booktitle={European conference on computer vision},
  pages={146--162},
  year={2022},
  organization={Springer}
}

@inproceedings{lerner2022viquae,
  title={ViQuAE, a dataset for knowledge-based visual question answering about named entities},
  author={Lerner, Paul and Ferret, Olivier and Guinaudeau, Camille and Le Borgne, Herv{\'e} and Besan{\c{c}}on, Romaric and Moreno, Jos{\'e} G and Lov{\'o}n Melgarejo, Jes{\'u}s},
  booktitle={Proceedings of the 45th International ACM SIGIR Conference on Research and Development in Information Retrieval},
  pages={3108--3120},
  year={2022}
}

@article{chen2023can,
  title={Can pre-trained vision and language models answer visual information-seeking questions?},
  author={Chen, Yang and Hu, Hexiang and Luan, Yi and Sun, Haitian and Changpinyo, Soravit and Ritter, Alan and Chang, Ming-Wei},
  journal={arXiv preprint arXiv:2302.11713},
  year={2023}
}

@inproceedings{mensink2023encyclopedic,
  title={Encyclopedic vqa: Visual questions about detailed properties of fine-grained categories},
  author={Mensink, Thomas and Uijlings, Jasper and Castrejon, Lluis and Goel, Arushi and Cadar, Felipe and Zhou, Howard and Sha, Fei and Araujo, Andr{\'e} and Ferrari, Vittorio},
  booktitle={Proceedings of the IEEE/CVF International Conference on Computer Vision},
  pages={3113--3124},
  year={2023}
}

@article{yan2024echosight,
  title={EchoSight: Advancing visual-language models with Wiki knowledge},
  author={Yan, Yibin and Xie, Weidi},
  journal={arXiv preprint arXiv:2407.12735},
  year={2024}
}

@article{li2024benchmarking,
  title={Benchmarking multimodal retrieval augmented generation with dynamic vqa dataset and self-adaptive planning agent},
  author={Li, Yangning and Li, Yinghui and Wang, Xinyu and Jiang, Yong and Zhang, Zhen and Zheng, Xinran and Wang, Hui and Zheng, Hai-Tao and Yu, Philip S and Huang, Fei and others},
  journal={arXiv preprint arXiv:2411.02937},
  year={2024}
}

@article{lee2024nv,
  title={NV-Embed: Improved Techniques for Training LLMs as Generalist Embedding Models},
  author={Lee, Chankyu and Roy, Rajarshi and Xu, Mengyao and Raiman, Jonathan and Shoeybi, Mohammad and Catanzaro, Bryan and Ping, Wei},
  journal={arXiv preprint arXiv:2405.17428},
  year={2024}
}

@article{sheng2024hybridflow,
  title   = {HybridFlow: A Flexible and Efficient RLHF Framework},
  author  = {Guangming Sheng and Chi Zhang and Zilingfeng Ye and Xibin Wu and Wang Zhang and Ru Zhang and Yanghua Peng and Haibin Lin and Chuan Wu},
  year    = {2024},
  journal = {arXiv preprint arXiv: 2409.19256}
}

@inproceedings{radford2021learning,
  title={Learning transferable visual models from natural language supervision},
  author={Radford, Alec and Kim, Jong Wook and Hallacy, Chris and Ramesh, Aditya and Goh, Gabriel and Agarwal, Sandhini and Sastry, Girish and Askell, Amanda and Mishkin, Pamela and Clark, Jack and others},
  booktitle={International conference on machine learning},
  pages={8748--8763},
  year={2021},
  organization={PMLR}
}

@article{wang2022text,
  title={Text embeddings by weakly-supervised contrastive pre-training},
  author={Wang, Liang and Yang, Nan and Huang, Xiaolong and Jiao, Binxing and Yang, Linjun and Jiang, Daxin and Majumder, Rangan and Wei, Furu},
  journal={arXiv preprint arXiv:2212.03533},
  year={2022}
}

@inproceedings{caffagni2024wiki,
  title={Wiki-llava: Hierarchical retrieval-augmented generation for multimodal llms},
  author={Caffagni, Davide and Cocchi, Federico and Moratelli, Nicholas and Sarto, Sara and Cornia, Marcella and Baraldi, Lorenzo and Cucchiara, Rita},
  booktitle={Proceedings of the IEEE/CVF Conference on Computer Vision and Pattern Recognition},
  pages={1818--1826},
  year={2024}
}

@article{chen2025learning,
  title={Learning to reason with search for llms via reinforcement learning},
  author={Chen, Mingyang and Li, Tianpeng and Sun, Haoze and Zhou, Yijie and Zhu, Chenzheng and Wang, Haofen and Pan, Jeff Z and Zhang, Wen and Chen, Huajun and Yang, Fan and others},
  journal={arXiv preprint arXiv:2503.19470},
  year={2025}
}

@article{schulman2017proximal,
  title={Proximal policy optimization algorithms},
  author={Schulman, John and Wolski, Filip and Dhariwal, Prafulla and Radford, Alec and Klimov, Oleg},
  journal={arXiv preprint arXiv:1707.06347},
  year={2017}
}

@article{shao2024deepseekmath,
  title={Deepseekmath: Pushing the limits of mathematical reasoning in open language models},
  author={Shao, Zhihong and Wang, Peiyi and Zhu, Qihao and Xu, Runxin and Song, Junxiao and Bi, Xiao and Zhang, Haowei and Zhang, Mingchuan and Li, YK and Wu, Yang and others},
  journal={arXiv preprint arXiv:2402.03300},
  year={2024}
}

@article{yu2025dapo,
  title={Dapo: An open-source llm reinforcement learning system at scale},
  author={Yu, Qiying and Zhang, Zheng and Zhu, Ruofei and Yuan, Yufeng and Zuo, Xiaochen and Yue, Yu and Dai, Weinan and Fan, Tiantian and Liu, Gaohong and Liu, Lingjun and others},
  journal={arXiv preprint arXiv:2503.14476},
  year={2025}
}

\newpage

\section{Supplementary Materials}

This section provides the detailed prompts used to construct the MMHops dataset. The following prompts were carefully engineered for state-of-the-art large language models like GPT-4o. 
They incorporate specific constraints and few-shot examples to ensure high-quality, structured output for both Bridging and Comparison question types.

\subsection{Bridging Question}

\subsubsection{Prompt for Sub-Question Generation}
Bridging questions are generated by creating a new sub-question based on a bridging entity. The following prompt template was used for this task.

\begin{promptbox}[Prompt: Bridging Sub-question Generation]
\textbf{Goal}

Given a Wikipedia entity and its corresponding Wikipedia content, you need to formulate a question about the entity or its attributes, provide the answer, and indicate the source sentence for the answer.

\textbf{Limitations}

\begin{itemize}
    \item \textbf{Question}:
    \begin{enumerate}
        \item The question must be about the given entity, inquiring about its attributes, such as the birth date of a person or the land area of a country.
        \item The original name of the given entity must appear in the question and cannot be a variant. It should only appear once; if it needs to appear multiple times, pronouns should be used.
        \item The question must not introduce additional information that allows the answer to be inferred from perspectives other than the given entity, including:
        \begin{enumerate}
            \item When asking about the given entity, do not focus on another independent entity in such a way that the question could be answered by only considering the other entity.
            \item Avoid introducing highly restrictive or directly answerable expressions.
        \end{enumerate}
        \item The question must unambiguously point to a unique answer. If the question is time-sensitive, a time frame should be included.
    \end{enumerate}

    \item \textbf{Answer}:
    \begin{enumerate}
        \item The answer should not be open-ended; it should be a specific noun, numerical value, or exact date, expressed in one word or a short phrase.
        \item Avoid questions that lead to incomplete answers or have multiple valid answers.
        \item The answer must be unambiguously derivable from the evidence sentence.
    \end{enumerate}

    \item \textbf{Evidence Sentence}:
    \begin{enumerate}
        \item The evidence sentence must form a complete reasoning chain. Both the subject of the question and the answer must appear in the evidence sentence.
    \end{enumerate}
\end{itemize}

\promptrule
\textbf{Examples}

\textbf{Example 1:}
\begin{itemize}[label=, leftmargin=0pt] 
    \item \textbf{Wikipedia entity:} Bean
    \item \textbf{Wikipedia content:} Taxonomy.The Fabaceae are placed in the order Fabales... The family now includes six subfamilies:Cercidoideae: 12 genera and ~335 species...
    \item \textbf{Output:}
    \begin{enumerate}
        \item Question: How many subfamilies are included in the Fabaceae family?
        \item Answer: Six
        \item Evidence sentence: The family now includes six subfamilies:Cercidoideae: 12 genera and ~335 species.
    \end{enumerate}
\end{itemize}

\textbf{Example 2:}
\begin{itemize}[label=, leftmargin=0pt]
    \item \textbf{Wikipedia entity:} Russia
    \item \textbf{Wikipedia content:} Culture.Russian culture has been formed by the nation's history... Russia is home to 30 UNESCO World Heritage Sites, 19 out of which are cultural...
    \item \textbf{Output:}
    \begin{enumerate}
        \item Question: How many UNESCO World Heritage Sites are there in Russia?
        \item Answer: 30
        \item Evidence sentence: Russia is home to 30 UNESCO World Heritage Sites, 19 out of which are cultural; while 27 more sites lie on the tentative list.
    \end{enumerate}
\end{itemize}

\promptrule
\textbf{Real Data}

\begin{itemize}[label=, leftmargin=0pt]
    \item \textbf{Wikipedia entity:} \{ENTITY\_ NAME\}
    \item \textbf{Wikipedia content:} \{WIKIPEDIA\_CONTENT\}
\end{itemize}

\textbf{Output Format}

You need to strictly follow the format below for output, and do not output anything other than the specified content. Only generate one question and answer pair:

\begin{enumerate}
    \item Question: [Your generated question]
    \item Answer: [Your generated answer]
    \item Evidence sentence: [The evidence sentence]
\end{enumerate}
\end{promptbox}

\subsubsection{Prompt for Question Quality Control}
After generating a sub-question, a quality control step is performed to ensure it does not contain information leakage and adheres to structural requirements. The following prompt was designed for this binary classification task.

\begin{promptbox}[Prompt: Question Quality Control]
\textbf{Goal}

Given a Wikipedia entity and a question about this entity, determine whether the question meets the following requirements. If it does, reply with \texttt{True}; if it does not, reply with \texttt{False}.

\textbf{Requirements}
\begin{enumerate}
    \item The question must focus on the given named entity itself and its attributes, not on derivatives of the entity. The entity name must appear in full and uninterrupted in the question.
    \item The question should not reveal additional information, such as restrictive descriptions, which would allow a model to bypass the given entity and answer the question.
    \item The question should not introduce independent new entities that allow it to be answered without using the given entity.
\end{enumerate}
\textbf{Guiding Principle:} The question must be unanswerable if the given entity name is removed from it.

\promptrule
\textbf{True Examples}

\textbf{Example 1:}
\begin{itemize}[label=, leftmargin=0pt]
    \item \textbf{Entity name:} Fabaceae
    \item \textbf{Question:} What type of fruits do Fabaceae produce after fertilization?
    \item \textbf{Output:} True
\end{itemize}

\textbf{Example 2:}
\begin{itemize}[label=, leftmargin=0pt]
    \item \textbf{Entity name:} Spain
    \item \textbf{Question:} How many World Heritage Sites does Spain have?
    \item \textbf{Output:} True
\end{itemize}

\textbf{Example 3:}
\begin{itemize}[label=, leftmargin=0pt]
    \item \textbf{Entity name:} Robert Adler
    \item \textbf{Question:} At what age did Robert Adler die?
    \item \textbf{Output:} True
\end{itemize}

\promptrule
\textbf{False Examples}

\textbf{Example 1:}
\begin{itemize}[label=, leftmargin=0pt]
    \item \textbf{Entity name:} AM General
    \item \textbf{Question:} In what year did AM General begin preliminary design work on the M998 Series High Mobility Multi-Purpose Wheeled Vehicle?
    \item \textbf{Explain:} The introduction of the "M998 Series..." allows one to obtain the answer by focusing only on this part, making it possible to skip the given entity "AM General."
    \item \textbf{Output:} False
\end{itemize}

\textbf{Example 2:}
\begin{itemize}[label=, leftmargin=0pt]
    \item \textbf{Entity name:} Inuit
    \item \textbf{Question:} What is the name of the northernmost city in the United States where the Inuit of Alaska live?
    \item \textbf{Explain:} The description "the northernmost city in the United States" allows the question to be answered directly without focusing on the entity "Inuit."
    \item \textbf{Output:} False
\end{itemize}

\textbf{Example 3:}
\begin{itemize}[label=, leftmargin=0pt]
    \item \textbf{Entity name:} Apple Inc.
    \item \textbf{Question:} When did Steve Jobs announce that Apple Computer, Inc. would be known as "Apple Inc."?
    \item \textbf{Explain:} "Steve Jobs" and "Apple Computer, Inc." can be used to infer the answer without needing to focus on the given entity "Apple Inc."
    \item \textbf{Output:} False
\end{itemize}

\promptrule
\textbf{Real Data}

\begin{itemize}[label=, leftmargin=0pt]
    \item \textbf{Entity name:} \{ENTITY\_NAME\}
    \item \textbf{Question:} \{QUESTION\_TEXT\}
\end{itemize}

\textbf{Output Format}

Please study the given examples and verify the provided real data. Output only \texttt{True} or \texttt{False} without any additional content.
\end{promptbox}


\subsubsection{Prompt for Question Merging}
To construct complex, multi-hop questions from simpler components, a merging step is essential. This prompt is designed to take an initial question and a follow-up question about its answer (the bridging entity) and combine them into a single, coherent multi-hop question.

\begin{promptbox}[Prompt: Question Merging]
\textbf{Goal}

Given an original question, its answer (which serves as a "bridging entity"), and a second question about that entity, merge the two questions into a single, new question.

\textbf{Requirements}
\begin{enumerate}
    \item \textbf{Substitution}: Replace the bridging entity in the second question with a descriptive clause derived from the original question. The name of the bridging entity must not appear in the final output.
    \item \textbf{Cohesion}: The output must be a single, grammatically complete question, not two separate clauses.
    \item \textbf{Fluency}: The merged question must be grammatically correct and sound natural in English.
\end{enumerate}

\promptrule
\textbf{Examples}

\textbf{Example 1:}
\begin{itemize}[label=, leftmargin=0pt]
    \item \textbf{Original question:} Who is the manufacturer of this vehicle?
    \item \textbf{Answer/Bridge entity:} AM General
    \item \textbf{Second question:} In what year did AM General begin producing purpose-built taxi-cabs?
    \item \textbf{Two-hop question:} In which year did the manufacturer of this vehicle begin producing purpose-built taxi-cabs?
\end{itemize}

\textbf{Example 2:}
\begin{itemize}[label=, leftmargin=0pt]
    \item \textbf{Original question:} What is this person's place of birth?
    \item \textbf{Answer/Bridge entity:} Brooklyn
    \item \textbf{Second question:} When did the Brooklyn Museum open?
    \item \textbf{Two-hop question:} When did the museum in this person's place of birth open?
\end{itemize}

\textbf{Example 3:}
\begin{itemize}[label=, leftmargin=0pt]
    \item \textbf{Original question:} Where is this bird native to?
    \item \textbf{Answer/Bridge entity:} New Zealand
    \item \textbf{Second question:} What percentage of New Zealand's indigenous vascular plants are endemic?
    \item \textbf{Two-hop question:} What percentage of indigenous vascular plants in the area where this bird is native are endemic?
\end{itemize}

\textbf{Example 4:}
\begin{itemize}[label=, leftmargin=0pt]
    \item \textbf{Original question:} What country does this building belong to?
    \item \textbf{Answer/Bridge entity:} Cambodia
    \item \textbf{Second question:} What is the temperature range in Cambodia in degrees Celsius?
    \item \textbf{Two-hop question:} What is the temperature range in the country to which this building belongs in degrees Celsius?
\end{itemize}

\promptrule
\textbf{Real Data}

\begin{itemize}[label=, leftmargin=0pt]
    \item \textbf{Original question:} \{ORIGINAL\_QUESTION\}
    \item \textbf{Answer/Bridge entity:} \{BRIDGE\_ENTITY\}
    \item \textbf{Second question:} \{SECOND\_QUESTION\}
\end{itemize}

\textbf{Output Format}

Please study the given examples and merge the two questions according to the requirements. Output only the final merged question, with no additional text or labels.
\begin{verbatim}
[Your generated multi-hop question]
\end{verbatim}
\end{promptbox}


\subsection{Comparison Question}

\subsubsection{Prompt for Comparison Question Generation}
This prompt is designed to generate high-quality comparative questions from two distinct text sources. It focuses on creating questions that require a precise, data-driven comparison between two entities, ensuring the output is structured, verifiable, and directly supported by the provided evidence.

\begin{promptbox}[Prompt: Comparison Question Generation]
\textbf{Goal}

Given two Wikipedia entities and paragraphs about these two entities, formulate a question comparing the two entities based on the content of the given paragraphs. Provide the answer and the sentence from which the evidence for the answer is derived.

\textbf{Requirements}
\begin{enumerate}
    \item The question must focus on comparing the same attribute of two given entities. It should not consider only a single entity, nor shift the comparison to other entities. The nature of the question must be comparative and not of any other type, such as summarization.
    \item The question should focus on comparing two entities with respect to specific attributes such as numerical values, dates, or years.
    \item The question description should be as detailed as possible, fully including the complete names of the two given entities (no splitting or abbreviations), and clearly specifying the expected answer format and units.
    \item The answer must be a single word or phrase, not a list of attributes for the two entities. Numerical answers should be expressed using Arabic numerals with units. The supporting evidence must be a complete sentence taken directly from the original text.
    \item If a valid question cannot be generated based on the given content, return False directly.
\end{enumerate}

\promptrule
\textbf{Examples}

\textbf{Example 1: Successful Numerical Comparison}
\begin{itemize}[label=, leftmargin=0pt]
    \item \textbf{Entity 1:} Gull-billed tern
    \item \textbf{Description:} This is a fairly large and powerful tern... It is 33 - 42 (cm) in length and 76 - 91 (cm) in wingspan. Body mass ranges from 150 - 292 (g)...
    \item \textbf{Entity 2:} Caspian tern
    \item \textbf{Description:} It is the world's largest tern with a length of 48–60 (cm), a wingspan of 127–145 (cm) and a weight of 530 - 782 (g)...
\end{itemize}
\textbf{Output:}
\begin{enumerate}
\item \textbf{Question}: What is the difference in centimeters between the maximum wingspans of the Gull-billed Tern and the Caspian Tern?
\item \textbf{Answer}: 54 cm
\item \textbf{Evidence sentences}:

Entity 1: It is 33 - 42 (cm) in length and 76 - 91 (cm) in wingspan.

Entity 2: It is the world's largest tern with a length of 48 – 60 (cm), a wingspan of 127 – 145 (cm) and a weight of 530 - 782 (g).

\end{enumerate}

\textbf{Example 2: Invalid Comparison}
\begin{itemize}[label=, leftmargin=0pt]
    \item \textbf{Entity 1:} Château de Termes
    \item \textbf{History:} ...the castle only fell to Simon de Montfort after a siege lasting four months, from August to November 1210...
    \item \textbf{Entity 2:} Château de Saissac
    \item \textbf{History:} Based on historical texts, it can be dated to at least 960...
\end{itemize}

\textit{Note: A valid comparison is not possible because the entities do not share a comparable attribute. One has a specific siege date, while the other has a date of origin.}

\textbf{Output:}
\begin{verbatim}
False
\end{verbatim}

\promptrule
\textbf{Real Data}

\begin{itemize}[label=, leftmargin=0pt]
    \item \textbf{Entity 1:} \{ENTITY\_1\_NAME\}
    \item \{ENTITY\_1\_DESCRIPTION\}
    \item \textbf{Entity 2:} \{ENTITY\_2\_NAME\}
    \item \{ENTITY\_2\_DESCRIPTION\}
\end{itemize}

\textbf{Output Format}

Strictly follow the format from the examples. Do not output any other content.
\begin{enumerate}
\item Question: ...
\item Answer: ...
\item Evidence sentences:...
\end{enumerate}
\textit{or}
\begin{verbatim}
False
\end{verbatim}
\end{promptbox}

\subsubsection{Prompts for Comparison Question Verification}
To ensure the quality of complex comparative questions, we employ a verification prompt that assesses them against strict logical and evidential criteria. This prompt validates the question's structure, the answer's format, and the verifiability of the claim against the provided evidence, as detailed below.

\begin{promptbox}[Prompt: Comparison Question Verification]
\textbf{Goal}

Given two Wikipedia entities, a question-answer pair, and an evidence sentence, verify whether the question-answer pair meets the following requirements. Reply with \texttt{True} if it does, otherwise \texttt{False}.

\textbf{Requirements}
\begin{enumerate}
    \item \textbf{Question Integrity:} The question must compare the two given entities, with both entity names fully appearing in the question. The question type must only compare numerical-related information for the given entities.
    \item \textbf{Answer Format:} The answer must be a numerical value with a unit, or a numerical range.
    \item \textbf{Verifiability:} The question must be clearly stated. Based on the supporting sentences, the question and answer must be verifiable without ambiguity. The attributes being compared in the question must be explicitly found in the supporting sentences; vague attribute references in the question do not meet this requirement.
\end{enumerate}

\promptrule
\textbf{Examples}

\textbf{Example 1:}
\begin{itemize}[label=, leftmargin=0pt]
\item \textbf{Entity 1:} Blanding's turtle
\item \textbf{Entity 2:} Northern map turtle
\item \textbf{Question:} Which turtle has a larger maximum clutch size, Blanding's turtle or Northern map turtle, and what is that maximum size?
\item \textbf{Answer:} Northern map turtle, 20 eggs
\item \textbf{Evidence Sentences:} Blanding's turtle: "The clutch size varies from region to region. In New York, the clutch size ranges from 5–12 eggs with an average of eight." \newline Northern map turtle: "The size of the clutch is between six and 20."
\item \textbf{Output:} \texttt{False}
\end{itemize}

\textbf{Example 2:}
\begin{itemize}[label=, leftmargin=0pt]
\item \textbf{Entity 1:} Glacier National Park (U.S.)
\item \textbf{Entity 2:} Yellowstone National Park
\item \textbf{Question:} Which national park has a greater number of documented mammal species, and by how many species?
\item \textbf{Answer:} Glacier National Park (U.S.) by 2 species
\item \textbf{Evidence Sentences:} Glacier National Park (U.S.): "Sixty-two species of mammals have been documented..." \newline Yellowstone National Park: "There are almost 60 species of mammals in the park..."
\item \textbf{Output:} \texttt{False}
\end{itemize}

\textbf{Example 3:}
\begin{itemize}[label=, leftmargin=0pt]
\item \textbf{Entity 1:} Humvee
\item \textbf{Entity 2:} Hummer H1
\item \textbf{Question:} What is the difference in years between the start of production of the Humvee and the Hummer H1?
\item \textbf{Answer:} 9 years
\item \textbf{Evidence Sentences:} Humvee: "AM General was awarded an initial contract in 1983 for 2,334 vehicles, the first batch of a five-year contract that would see 55,000 vehicles delivered to the U.S. military." \newline Hummer H1: "Originally released in the civilian market March 14, 1992, the Hummer H1 became known from photographs published during Operation Desert Storm and the enthusiastic promotion by actor Arnold Schwarzenegger."
\item \textbf{Output:} \texttt{False} \newline The evidence is not sufficient and direct enough.
\end{itemize}

\textbf{Example 4:}
\begin{itemize}[label=, leftmargin=0pt]
\item \textbf{Entity 1:} Crescent City Connection
\item \textbf{Entity 2:} Blue Water Bridge
\item \textbf{Question:} What is the difference in years between the opening of the first spans of the Crescent City Connection and the Blue Water Bridge?
\item \textbf{Answer:} 20 years
\item \textbf{Evidence Sentences:} Crescent City Connection: "Construction of the first span began in November 1954, and the bridge opened in April 1958 as the Greater New Orleans Bridge." \newline Blue Water Bridge: "The first bridge was fully opened to traffic on October 10, 1938."
\item \textbf{Output:} \texttt{True}
\end{itemize}

\textbf{Example 5:}
\begin{itemize}[label=, leftmargin=0pt]
\item \textbf{Entity 1:} Rumex obtusifolius
\item \textbf{Entity 2:} Rumex crispus
\item \textbf{Question:} What is the difference in centimeters between the maximum heights of Rumex obtusifolius and Rumex crispus?
\item \textbf{Answer:} 50 cm
\item \textbf{Evidence Sentences:} 

Rumex obtusifolius: "Rumex obtusifolius is a perennial herbaceous flowering plant that grows to a height of 40 to 150 (cm)." 

Rumex crispus: "The plant produces an inflorescence or flower stalk that grows to 1.5 (m) high."

\item \textbf{Output:} \texttt{False} 

The answer contains factual errors and inaccurate calculation results.
\end{itemize}

\promptrule
\textbf{Real Data}

\begin{itemize}[label=, leftmargin=0pt]
\item \textbf{Entity 1:} \{ENTITY\_1\}
\item \textbf{Entity 2:} \{ENTITY\_2\}
\item \textbf{Question:} \{QUESTION\}
\item \textbf{Answer:} \{ANSWER\}
\item \textbf{Evidence Sentences:} \{EVIDENCE\_SENTENCE\}
\end{itemize}

\textbf{Output Format}

Strictly follow the format from the examples. The output must be the single word \texttt{True} or \texttt{False}.
\end{promptbox}

\subsubsection{Prompts for Comparison Question Rewriting}
To adapt our comparative questions for multimodal contexts, we use a prompt to rewrite them into a format that references images instead of named entities. This process instructs the LLM to replace entity names with generic references like "the building in the first image" while preserving the original question's comparative intent, as detailed below.

\begin{promptbox}[Prompt: Comparison Question Rewriting]
\textbf{Goal}

Given a question that compares two entities, your task is to rewrite the question to eliminate the entity names, replacing them with references to two images, while keeping the question's meaning unchanged. Your response should be the rewritten question.

\textbf{Requirements}
\begin{enumerate}
    \item \textbf{Preserve Meaning:} The rewritten question must not include the original entity names, but its meaning must remain the same and still point to the given answer. To ensure the answer format remains as entity names, add "(answer the name)" where appropriate.
    \item \textbf{Image Referencing:} The two entities correspond to two images: entity 1 is in the first image, and entity 2 is in the second. Replace references to the entities in the question with references to the images, either individually or collectively.
    \item \textbf{Use Entity Type:} In the rewritten question, you can use the type of the entity (e.g., car, airplane, animal, plant, building) to replace the entity name for clarity.
    \item \textbf{Clarity and Brevity:} The rewritten question should be concise and natural, avoiding complex or awkward phrasing.
\end{enumerate}

\promptrule
\textbf{Examples}

\textbf{Example 1:}
\begin{itemize}[label=, leftmargin=0pt]
\item \textbf{Entity 1:} Common toad
\item \textbf{Entity 2:} Common frog
\item \textbf{Question:} How many more eggs can a Common Toad lay compared to a Common Frog at maximum capacity in eggs?
\item \textbf{Answer:} 30,000 eggs
\item \textbf{Output:} How many more eggs can the animal in the first image lay compared to the second image at maximum capacity in eggs?
\end{itemize}

\textbf{Example 2:}
\begin{itemize}[label=, leftmargin=0pt]
\item \textbf{Entity 1:} Trumpetfish
\item \textbf{Entity 2:} Cornetfish
\item \textbf{Question:} What is the difference in meters between the maximum lengths of the Trumpetfish and the Cornetfish?
\item \textbf{Answer:} 1 m
\item \textbf{Output:} What is the difference in meters between the maximum lengths of the fish in the first image and the second image?
\end{itemize}

\promptrule
\textbf{Real Data}

\begin{itemize}[label=, leftmargin=0pt]
\item \textbf{Entity 1:} \{ENTITY\_1\}
\item \textbf{Entity 2:} \{ENTITY\_2\}
\item \textbf{Question:} \{QUESTION\}
\item \textbf{Answer:} \{ANSWER\}
\end{itemize}

\textbf{Output Format}

The output should be only the rewritten question.
\end{promptbox}

\subsection{Answer Processing and Normalization}

\subsubsection{Prompt for Answer Verification and Classification}
To ensure the generated question-answer pairs are valid and to categorize them for analysis, a verification and classification prompt is used. This step confirms that the answer is derivable from the evidence and assigns it a specific data type.

\begin{promptbox}[Prompt: Answer Verification and Classification]
\textbf{Goal}

Given a question-answer pair and the corresponding evidence sentence, verify whether the pair meets the requirements and classify the type of the answer.

\textbf{Requirements}
\begin{enumerate}
    \item The answer can be precisely inferred from the evidence sentence.
    \item The answer must be a specific number, numerical range, year, date, or string. It cannot be a vague concept.
\end{enumerate}

\textbf{Classification}

Based on the answer, classify it into one of the following three types. If it does not meet the requirements or is a vague concept, classify it as \texttt{False}.
\begin{enumerate}
    \item For date-type answers, respond with \textbf{Time}. This is limited to specific years (e.g., 1897) and dates (e.g., 1 January 1981). General ranges like "the 16th century" are classified as \texttt{String}.
    \item For numerical-type answers, respond with \textbf{Numerical}, which includes specific numbers or specific numerical ranges. Answers of this type generally include a unit of measurement; if no unit is provided, the answer should represent a count or percentage.
    \item For string-type answers, respond with \textbf{String}. This applies to any other specific, non-numerical, non-date answer (e.g., Zebedee, 4th, 16th century).
\end{enumerate}
For answers that do not meet the above requirements, cannot be inferred from the given evidence sentence, or are vague concepts or ranges, respond with \textbf{False}.

\promptrule
\textbf{Time Examples}

\textbf{Example 1:}
\begin{itemize}[label=, leftmargin=0pt]
    \item \textbf{Question:} When did the Brooklyn Museum open?
    \item \textbf{Answer:} 1897
    \item \textbf{Evidence Sentences:} The Brooklyn Museum, opened in 1897, is New York City’s second-largest public art museum.
    \item \textbf{Output:} Time
\end{itemize}

\promptrule
\textbf{Numerical Examples}

\textbf{Example 1:}
\begin{itemize}[label=, leftmargin=0pt]
    \item \textbf{Question:} What percentage of all genera of land plants were angiosperms in the Maastrichtian?
    \item \textbf{Answer:} 50\% to 80\%
    \item \textbf{Evidence Sentences:} From 50\% to 80\% of all genera of land plants were angiosperms...
    \item \textbf{Output:} Numerical
\end{itemize}

\textbf{Example 2:}
\begin{itemize}[label=, leftmargin=0pt]
    \item \textbf{Question:} How many living subdivisions does Cryptobranchoidea have?
    \item \textbf{Answer:} Two
    \item \textbf{Evidence Sentences:} It has two living subdivisions, Cryptobranchidae... and Hynobiidae...
    \item \textbf{Output:} Numerical
\end{itemize}

\promptrule
\textbf{String Examples}

\textbf{Example 1:}
\begin{itemize}[label=, leftmargin=0pt]
    \item \textbf{Question:} Who was John the Apostle's father?
    \item \textbf{Answer:} Zebedee
    \item \textbf{Evidence Sentences:} John the Apostle was the son of Zebedee and the younger brother of James the Great.
    \item \textbf{Output:} String
\end{itemize}

\textbf{Example 2:}
\begin{itemize}[label=, leftmargin=0pt]
    \item \textbf{Question:} When did the dominance of Gothic architecture begin to wane?
    \item \textbf{Answer:} 16th century
    \item \textbf{Evidence Sentences:} Beginning in the 16th century... the dominance of Gothic architecture began to wane.
    \item \textbf{Output:} String
\end{itemize}

\promptrule
\textbf{False Examples}

\textbf{Example 1:}
\begin{itemize}[label=, leftmargin=0pt]
    \item \textbf{Question:} How many members of parliament are elected in Turkey?
    \item \textbf{Answer:} Five to six hundred thousand
    \item \textbf{Evidence Sentences:} Out of a population of 9.5 million, it is estimated that five to six hundred thousand people sing in choirs.
    \item \textbf{Explain:} The evidence sentence does not support the question, and the answer "five to six hundred thousand" is presented as a vague estimate, not a precise range.
    \item \textbf{Output:} False
\end{itemize}

\promptrule
\textbf{Real Data}

\begin{itemize}[label=, leftmargin=0pt]
    \item \textbf{Question:} \{QUESTION\}
    \item \textbf{Answer:} \{ANSWER\}
    \item \textbf{Evidence Sentences:} \{EVIDENCE\_SENTENCE\}
\end{itemize}

\textbf{Output Format}

Please classify the given real data according to the instructions. Output only one of the specified classification types: \texttt{Time}, \texttt{Numerical}, \texttt{String}, or \texttt{False}.
\end{promptbox}


\subsubsection{Prompt for Numerical Answer Normalization}
To create a robust and machine-readable evaluation set for questions with numerical answers, the raw text answers must be standardized. This prompt automates the process of validating, extracting, and normalizing numerical data, and refines the corresponding question to ensure clarity for automated evaluation.

\begin{promptbox}[Prompt: Numerical Answer Normalization]
\textbf{Goal}

Given a question-answer pair and its evidence sentence, standardize the answer into a machine-readable format. This involves extracting the numerical value and its unit, and rewriting the question to explicitly request that unit.

\textbf{Processing Steps}
\begin{enumerate}
    \item \textbf{Extract \& Standardize Value:} Extract the numerical value(s), converting words (e.g., "Seven") to digits (e.g., "7"). Remove all units and formatting (e.g., commas, symbols). For a range, list the lower and upper bounds. The result should be a list of numbers.
    \item \textbf{Identify Unit:} Extract the unit of measurement (e.g., \%, million, °C). If the answer is a simple count, the unit is \texttt{None}.
    \item \textbf{Rewrite Question:} Modify the original question, if necessary, to ensure it explicitly asks for the answer in the identified unit (e.g., "What is the population?" becomes "What is the population in millions?").
\end{enumerate}

\promptrule
\textbf{Examples}

\textbf{Example 1:}
\begin{itemize}[label=, leftmargin=0pt]
    \item \textbf{Question:} What percentage of Germany's health care system was government-funded according to the World Health Organization?
    \item \textbf{Answer:} 77\%
    \item \textbf{Evidence Sentences:} According to the World Health Organization (WHO), Germany's health care system was 77\% government-funded and 23\% privately funded.
    \item \textbf{Output:}
\begin{enumerate}
\item Value: [77]
\item Unit: %
\item Question: What percentage of Germany's health care system was government-funded according to the World Health Organization?
\end{enumerate}
\end{itemize}

\textbf{Example 2:}
\begin{itemize}[label=, leftmargin=0pt]
    \item \textbf{Question:} What is the population of Denmark as of 2022?
    \item \textbf{Answer:} 5.91 million
    \item \textbf{Evidence Sentences:} As of 2022, it had a population of 5.91 million (1 August 2022)...
    \item \textbf{Output:}
\begin{enumerate}
\item Value: [5.91]
\item Unit: million
\item Question: What is the population of Denmark in millions as of 2022?
\end{enumerate}
\end{itemize}

\textbf{Example 3:}
\begin{itemize}[label=, leftmargin=0pt]
    \item \textbf{Question:} What is the temperature range in Cambodia?
    \item \textbf{Answer:} 21 to 35 (°C)
    \item \textbf{Evidence Sentences:} Cambodia has a temperature range from 21 to 35 (°C) and experiences tropical monsoons.
    \item \textbf{Output:}
\begin{enumerate}
\item Value: [21, 35]
\item Unit: °C
\item Question: What is the temperature range in Cambodia in degrees Celsius?
\end{enumerate}
\end{itemize}

\promptrule
\textbf{Real Data}

\begin{itemize}[label=, leftmargin=0pt]
    \item \textbf{Question:} \{QUESTION\}
    \item \textbf{Answer:} \{ANSWER\}
    \item \textbf{Evidence Sentences:} \{EVIDENCE\_SENTENCE\}
\end{itemize}

\textbf{Output Format}

Strictly follow the format from the examples. The output must be either the three-line structure or the single word \texttt{False}. Do not output any other content.
\begin{enumerate}
\item Value: [...]
\item Unit: ...
\item Question: ...
\end{enumerate}
\textit{or}
False
\end{promptbox}


\subsubsection{Prompt for String Answer Normalization}
To robustly evaluate answers that are strings (e.g., names, places, concepts), it is necessary to account for all valid, synonymous expressions present in the source text. This prompt standardizes string answers by generating a comprehensive list of all equivalent terms found in the evidence.

\begin{promptbox}[Prompt: String Answer Normalization]
\textbf{Goal}

Given a question-answer pair and the corresponding evidence sentence, standardize the answer, write out all synonymous expressions of the answer, and generate an answer list for evaluation.

\textbf{Processing Steps}
\begin{enumerate}
    \item \textbf{Validate Answer:} First, verify that the answer is a string and can be fully and accurately derived from the provided evidence sentence. If either condition is not met, the output must be \texttt{False}.
    \item \textbf{Generate Synonym List:} If the answer is valid, identify and extract all equivalent expressions, names, or abbreviations for the answer from the evidence. Collate these into a single list of strings. All strings in the list should be in lowercase. If no synonyms are found, the list should contain only the original answer in lowercase.
\end{enumerate}

\promptrule
\textbf{Real Data}

\begin{itemize}[label=, leftmargin=0pt]
    \item \textbf{Question:} \{QUESTION\}
    \item \textbf{Answer:} \{ANSWER\}
    \item \textbf{Evidence Sentences:} \{EVIDENCE\_SENTENCE\}
\end{itemize}

\textbf{Output Format}

Strictly follow the format from the examples. The output must be either the \texttt{Answer\_eval} list or the single word \texttt{False}. Do not output any other content.

Answer\_eval: [...]

\textit{or}

False

\end{promptbox}

\subsubsection{Prompts for Time-based Answer Normalization}
For time-based answers, this process involves standardizing various formats into a canonical form and generating an extensive list of acceptable variants for flexible matching, as detailed in the prompt below.

\begin{promptbox}[Prompt: Time-based Answer Normalization]
\textbf{Goal}

Given a question-answer pair and the corresponding evidence sentence, standardize the answer and generate an evaluation list for the answer. Create the following two contents: 
\texttt{Answer} and \texttt{Answer\_eval}.

\textbf{Processing Steps}
\begin{enumerate}
    \item Check if the answer is a specific year or a specific date. If it is, proceed to the next steps; if not, return \textbf{False}.
    \item Verify if the answer can be inferred from the evidence sentence. If it can, fill the "answer" list with the given answer, including only the year or date without any units.
    \item  Generate Answer\_eval: 
    \begin{itemize}
        \item  If the answer is a specific year, include the exact year in the ''Answer\_eval'' list and add a range of plus and minus one year to create an evaluation list. 
        \item If the answer is a specific date that includes a year, first include all standard expressions of the date in the ''Answer\_eval'' list. Then, adjust the year by plus and minus one and add all possible expressions for these adjusted dates. Finally, remove the date and list all possible years. 
        \item If the answer is a specific date that does not include a year, list all possible standard expressions of the date in the ''Answer\_eval'' list.
    \end{itemize}
    \item If the above steps cannot be completed, return false directly.
\end{enumerate}

\promptrule
\textbf{Examples}

\textbf{Example 1:}
\begin{itemize}[label=, leftmargin=0pt]
\item \textbf{Question:} When did the Brooklyn Museum open?
\item \textbf{Answer:} 1897
\item \textbf{Evidence Sentences:} The Brooklyn Museum, opened in 1897, is New York City’s second-largest public art museum.
\item \textbf{Output:}
\begin{enumerate}
\item \texttt{Answer: ["1897"]}
\item \texttt{Answer\_eval: ["1897", "1896", "1898"]}
\end{enumerate}
\end{itemize}

\textbf{Example 2:}
\begin{itemize}[label=, leftmargin=0pt]
\item \textbf{Question:} When did Greece become the tenth member of the European Communities?
\item \textbf{Answer:} 1 January 1981
\item \textbf{Evidence Sentences:} Greece became the tenth member of the European Communities (subsequently subsumed by the European Union) on 1 January 1981, ushering in a period of sustained growth.
\item \textbf{Output:}
\begin{enumerate}
\item \texttt{Answer: ["1 January 1981"]}
\item \texttt{Answer\_eval: ["1 January 1981", "January 1 1981", "1981 January 1", "1 1 1981", "1981 1 1", "1981", "1 January 1980", "January 1 1980", "1980 January 1", "1 1 1980", "1980 1 1", "1980", "1 January 1982", "January 1 1982", "1982 January 1", "1 1 1982", "1982 1 1", "1982"]}
\end{enumerate}
\end{itemize}

\textbf{Example 3:}
\begin{itemize}[label=, leftmargin=0pt]
\item \textbf{Question:} When is Russia's Unity Day celebrated?
\item \textbf{Answer:} 4 November
\item \textbf{Evidence Sentences:} Unity Day on 4 November, commemorating the 1612 uprising which marked the end of the Polish occupation of Moscow.
\item \textbf{Output:}
\begin{enumerate}
\item \texttt{Answer: ["4 November"]}
\item \texttt{Answer\_eval: ["4 November", "November 4", "4th November", "November 4th", "4 11", "11 4"]}
\end{enumerate}
\end{itemize}

\textbf{Example 4:}
\begin{itemize}[label=, leftmargin=0pt]
\item \textbf{Question:} When did Denmark stop issuing new licences for oil and gas extraction?
\item \textbf{Answer:} December 2020
\item \textbf{Evidence Sentences:} Denmark stopped issuing new licences for oil and gas extraction in December 2020.
\item \textbf{Output:}
\begin{enumerate}
\item \texttt{Answer: ["December 2020"]}
\item \texttt{Answer\_eval: ["December 2020", "2020 December", "12 2020", "2020 12", "2020", "December 2021", "2021 December", "12 2021", "2021 12", "2021", "December 2019", "2019 December", "12 2019", "2019 12", "2019"]}
\end{enumerate}
\end{itemize}

\promptrule
\textbf{Real Data}

\begin{itemize}[label=, leftmargin=0pt]
\item \textbf{Question:} \{QUESTION\}
\item \textbf{Answer:} \{ANSWER\}
\item \textbf{Evidence Sentences:} \{EVIDENCE\_SENTENCE\}
\end{itemize}

\textbf{Output Format}

Strictly follow the format from the examples. The output must be either the two-line structure or the single word \texttt{False}. Do not output any other content.
\begin{enumerate}
\item \texttt{Answer: [...]}
\item \texttt{Answer\_eval: [...]}
\end{enumerate}
\textit{or}
\texttt{False}
\end{promptbox}

\end{document}